# Semi-Supervised Recognition under a Noisy and Fine-grained Dataset

Cheng Cui*, Zhi Ye*, Yangxi Li*, Xinjian Li*,
MinYang*, Kai Wei *, Bing Dai*, Yanmei Zhao*, Zhongji Liu*, Rong Pang*
National Internet Emergency Center(CNCERT), Baidu
{laxlyezhi, weikai105b, cuicheng0101, yangminbupt}@gmail.com

## Abstract

*Simi-Supervised Recognition Challenge-FGVC7 is a challenging fine-grained recognition competition. One of the difficulties of this competition is how to use unlabeled data. We adopted pseudo-tag data mining to increase the amount of training data. The other one is how to identify similar birds with a very small difference, especially those have a relatively tiny main-body in examples. We combined generic image recognition and fine-grained image recognition method to solve the problem. All generic image recognition models were training using PaddleClas[1] . Using the combination of two different ways of deep recognition models, we finally won the third place in the competition.*

## 1. Introduction

Simi-Supervised Recognition Challenge-FGVC7 focused on learning from partially labeled data, which is a form of semi-supervised learning. The dataset is designed to expose some of the challenges encountered in a realistic setting, such as the fine-grained similarity between classes, significant class imbalance, and domain mismatch between the labeled and unlabeled data. This challenge is part of the FGVC7 workshop in CVPR 2020.

The data set is divided into the following three parts, the first is labeled data, a total of 5959 pictures, the second is unlabeled data whose categories belong to the labeled data, a total of 26640 pictures, and the third is unlabeled data but its category is out of the labeled data, a total of 122,208 pictures. The labeled data is divided into a training set of 3959 pictures and a validation set of 2000 pictures. The training data is unbalanced, while the validation data balanced. In addition, there is a total of 8,000 pictures in the test set, including 4,000 public data and 4,000 private data.

In the traditional fine-grained recognition tasks, most common fine-grained recognition methods are used. How-ever, in this competition, fine-grained method can not solve this problem well due to the small amount of training data and small targets with complex backgrounds in examples. In addition to the fine-grained method, a lot of generic identification models were adopted. And we used PaddleClas to train these models, mainly for the reasons as follows:

(1) PaddleClas is a toolset for image classification tasks prepared for the industry and academia written in PaddlePaddle[2]. It helps users train better computer vision models and apply them in real scenarios.

(2) Based on Image-Net1k dataset, PaddleClas provides 23 series of image classification networks such as ResNet, ResNet_vd, Res2Net, HRNet, and MobileNetV3

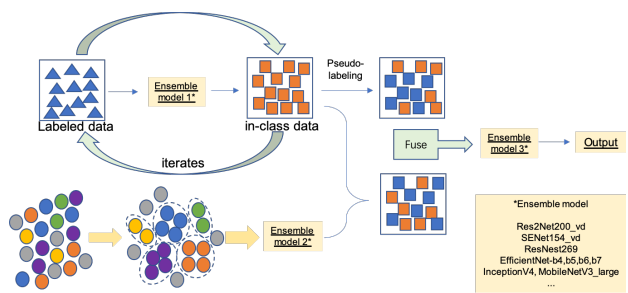

Figure 1. Diagram of the overall training framework. For the in-class dataset (labeled and unlabeled), we trained the labeled data using basic classification and fine-grained classification algorithms with various optimization methods, the different outputs were merged together to do the pseudo-labeling on in-class dataset. The processing pipeline we called ensemble modeling as shown in Figure 2. The pseudo-labeling process iterates several times to enlarge the pseudo-label samples from in-class set gradually. To effectively utilize the out-of-class dataset, we clustered the data into 10k classes then train a classification model for a good pretraining start. The model was finetuned with labeled data and then used for mining in-class dataset as before. Finally we fused two pseudo-labeling dataset to build it as our final training set. In practice, we tried different mining and fusion ways with minor parameter changes to get final results.

---

*These authors contributed equally to this work.
[1]The PaddleClas is available at https://github.com/PaddlePaddle/PaddleClas

[2]https://github.com/PaddlePaddle/Paddle

with brief introductions, reproduction configurations and training tricks.

(3) PaddleClas provides a Simple Semi-supervised Label Distillation method (SSLD). With this method, different models on Image-Net1k validation dataset have 3% absolute improvement(Top1 accuracy). For example, ResNet50 and ResNet101 achieved top-1 accuracy of 83% and 83.7%, respectively.

(4) PaddleClas provides the reproduction of the 8 data augmentation algorithms and the evaluation of the effect in a unified environment. We can use data augmentation such as Mixup, Cutmix, Cutout, RandomAugment, etc conveniently.

In this paper, we will introduce our solution to this competition. Firstly, we trained several specific deep models with the generic image recognition method. Secondly, these models were used to mine more examples from the in-class data through voting strategy. At the same time, we clustered the out-of-class unlabeled data to 10k class to produce a pretrained model on the bird data. Then, we used this pretrained model to train a new model on the train-validation dataset and the new model was used to fuse the mined data. The fused data was fixed as our final training set. Finally, we retrained a generic image recognition models and a fine-grained model with highest accuracy on this final training set, and merged the results of the two best models of different schemes.

## 2. Proposed Solution

We now describe our approach in detail, which consists of the follwing models and main steps. The whole pipline is show in Figure 1

### 2.1. Data selection

There are only 3959 labeled images provided, so the main difficulty lies in the utilization of unlabeled data. We've tried many ways to label in-class data and out-class data, including some semi-supervised methods: FixMatch [1] and MixMatch [2], but their results are not ideal. Take FixMatch as an example, we used out-of-class data as unlabeled data while merging the train-set, validation-set and in-class samples whose classification score larger than a specific threshold as labeled data. However the accuracy get lower than training with pseudo labeled in-class data. So we adopt the pseudo label method finally. Firstly, we used various basic classification and fine-grained classification algorithms to train labeled data, the details is described below. Secondly, we assemble models through the top-1 voting fusion and verify the accuracy on the validation set. Specifically, we select the model combination with the highest accuracy to label the in-class data. Lastly, we select the in-class data with high confidence and add them to the labeled data set. This whole process iterated several times until the accuracy of merged model converged.

We clustered the out-of-class 120k data into 10k classes. Then, we trained the models using the 10k-class data and the top1 accuracy reached 92%. We used this model as the pretrained model to train the train-validation dataset, which had a better performance than using ImageNet pretrained model. After the training, we used this model's prediction result on in-class data to intersect the result of the data mined by the ImageNet pretrained model training. The intersection part is the in-class training data set we finally choose.

### 2.2. Generic image recognition method

In terms of the selection of backbone, we found that the model with high accuracy in ImageNet will also have high accuracy in 3959 training sets and 2000 validation sets, so we followed this principle and selected the ImageNet pretrained model with high accuracy: SENet154_vd, Res2Net200_vd_26w_4s, DPN107, DPN131, HRNet_W64_C, ResNet200_vd, ResNeXt152_Vd_64x4d, InceptionV4, Xception65, DenseNet161, DenseNet264 [3, 4, 5, 6, 7, 8, 9], etc. The experimental results show that their merged results achieved good results. We conducted experiments on model training tricks on 3959 training sets and 2000 validation sets. Some conclusions are as follows:

(1) Using a complex model network (such as ResNet200_vd, ResNeXt152_vd_64x4d) is about 5~10 points higher than a simple network such as ResNet50_vd;

(2) In large scale dataset, epoch=50 is a more suitable parameter;

(3) Through grid-search, batch_size=128, learning_rate=0.025 is the best parameter combination;

(4) Cutmix improves about 2 points which is better than other data augmentation methods (such as mixup). If the data set is large, cutmix may be unhelpful;

(5) There is no obvious help using dropout. While using dropblock in the model will increase by about 1 point;

(6) Adjusting the training and validation crop_size from 224 to 448 will increase by about 4 points;

(7) Increasing model regularization can improve the accuracy. The epsilon value of label_smooth [10] is better to be 0.2 for small models such as ResNet50 and 0.3 for large models such as SE_ResNet152;

(8) In the model test phase, test time augmentation (TTA) is adopted. Through three data preprocessing methods: Resize+CenterCrop, Resize+RandomCrop, Resize+RandomCrop+RandomHorizontalFlip, each model can output 3 results, which will increase by 0.3~0.5 Points;

(9) During model testing, using the 144-crop method will increase it by about 1 point, but it takes more time.

(10) The 'Fix strategy' is very effective in postprocessing, which can increase by more than 1 point [11].

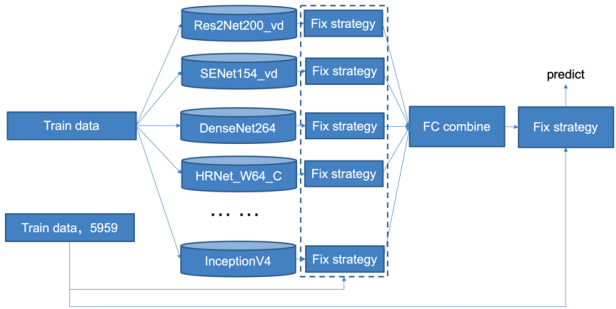

Figure 2. In the generic image recognition method , multiple generic recognition models are trained, and then the precision is further improved through 'Fix strategy'[11]. The model trained by 'Fix strategy' is fused, and the fused model is trained again by 'Fix strategy', in which the data of training 'Fix strategy' is 5959 train+val dataset.

Based on the above conclusions, we trained a total of 8 models and fused the results. Finally, as a result of the fusion, we reached 16.6% on the public list and 16.0% on the private list. The fusion strategy is mainly to add the weighted FC results before sotfmax, the weight is determined by its accuracy, and the higher the accuracy of the model, the greater the weight. At the same time, in the process of fusion, we also used the train-validation data of 5959 to perform the 'Fix strategy'[11]. The detailed operation is show in Figure 2

### 2.3. Fine-grained recognition method

In this competition, we tried many advanced image recognition methods, but after several iterations, the accuracy reached the bottleneck, and it's difficult to be improved. On the one hand, we think that the amount of data with labels is very small (3959), while the effect of general data augmentation methods will reach the ceiling soon, such as RandomCrop, RandomHorizontalFlip, ColorJitter and so on; on the other hand, we find that there are high intra-class variances and low inter-class variances by analyzing bad cases, which is a great challenge to the generic image recognition. At this time, we turn our attention to the fine-grained classification algorithm. We use Weakly Supervised Data Augmentation Network(WS-DAN) [12] and Destruction and Construction Learning for Fine-grained Image Recognition(DCL) [13] here.

With WSDAN, we implemented fine-grained data augmentation. Specifically, for each training image, we first generate attention maps to represent the object's discriminative parts by weakly supervised learning, then we augment the image under the guidance of these attention maps, including attention cropping and attention dropping. WS-DAN improves the classification accuracy in two aspects: first, images can be seen better since more discriminative parts' features will be extracted; then, attention regions provide accurate location of object, which ensures our model to look at the object closer and further improve the performance. Comprehensive experiments in common fine-grained visual classification datasets also prove the effectiveness of WS-DAN.

DCL is also a data augmentation method for fine-grained recognition, it enhances the difficulty of fine-grained recognition and exercises the classification model to acquire expert knowledge. Besides the standard classification backbone network, another "destruction and construction" stream is introduced to carefully "destruct" and then "reconstruct" the input image, for learning discriminative regions and features. More specifically, for "destruction", we first partition the input image into local regions and then shuffle them by a Region Confusion Mechanism (RCM). To correctly recognize these destructed images, the classification network has to pay more attention to discriminative regions for spotting the differences. For "construction", a region alignment network, which tries to restore the original spatial layout of local regions, is followed to model the semantic correlation among local regions, experimental results also show its effectiveness.

We further optimize WSDAN and DCL by the following methods:

(1) About tricks for weakly supervised learning, we find that label smooth, warmup + Learning rate with cosine decay and 144-crop prediction can improved the results. Besides, larger batch size and image crop size can increase by about 1.5 points;

(2) With the increase of iterations of inclass data, the number of trusted pseudo labels is increasing, which can significantly improve the model result by 10 points or more;

(3) We try to use different SOTA backbones to train WS-DAN, such as ResNeSt269 [14] and EfficientNet-B7 [15], the individual model output didn't show up any exciting result, but by combining these results, the accuracy can be further improved about 1.5 points.

### 2.4. The combination of generic image recognition and fine-grained recognition

The combination of generic image recognition method and fine-grained recognition pipeline is very important to improve the recognition effect of this challenge dataset. We train several basic generic classification models using PaddleClas toolbox from PaddlePaddle and specific fine-grained models with Pytorch framework respectively. We assembled various models' results by merging the last fc layer output before the softmax activatation and this operation will give us more robust classification results.

In the prediction phase, we first locate the target objects by our attention visualization method. Some visualization samples are shown in Figure 3. Then, general or fine-grained recognition model or combination of them is sclected to classify different samples under direction of the scale of targets. For example, for the very small targets, especially those hidden in the surrounding environment, or almost integrated with the environment, which makes our naked eyes almost unable to identify, generic image recognition will pay more attention to the environmental information in the image, which also plays a very good role in the correct recognition process. For the middle large objects, we give a higher confidence to the fine-grained image recognition model while for the very large objects, we give a higher confidence to the generic model.

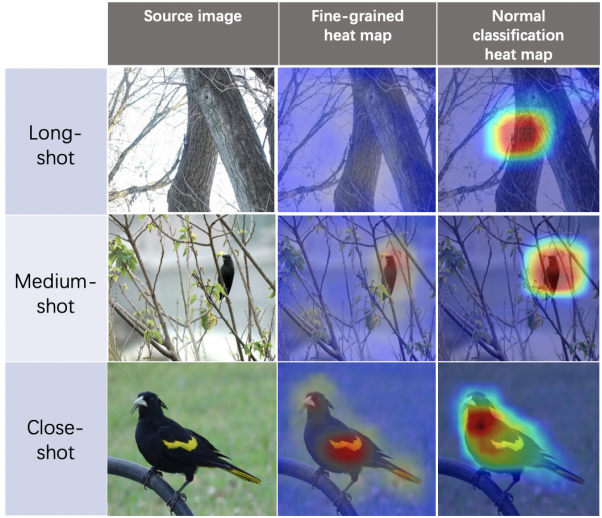

Figure 3. Visualization of attention region among long-shot, medium-shot and close-shot bird images. We found attention region area (especially fine-grained model) is helpful for us to judge the shoot type of the image. If it is a long-shot image, we give a higher confidence to the generic model, otherwise we give more confidence to fine-grained model when we do the model fusion.

| Model | Training resolution | Test resolution | Public | Private |
|---|---|---|---|---|
| SENet154_vd | 448x448 | 576x576 | —— | —— |
| Res2Net200_vd | 448x448 | 576x576 | 18.5% | 17.5% |
| DenseNet264 | 448x448 | 576x576 | —— | —— |
| DenseNet161 | 448x448 | 576x576 | —— | —— |
| ResNet200_vd | 448x448 | 576x576 | —— | —— |
| Inception_V4 | 448x448 | 576x576 | —— | —— |
| Xception65 | 448x448 | 576x576 | —— | —— |
| HRNet_W64_C | 448x448 | 576x576 | —— | —— |
| Combine_8_model | —— | 576x576 | 16.6% | 16.0% |

Table 1. Results of different generic image recognition models on the test dataset.

| model(WS-DAN) | Training resolution | Testing resolution | Public | Private |
|---|---|---|---|---|
| ResNet152 | 600x600 | 600x600 | 20.1% | 20.3% |
| Res2Net101 | 600x600 | 600x600 | 18.9% | 19.5% |
| SENet154 | 600x600 | 600x600 | —— | —— |
| Combine_3_model | —— | 600x600 | 16.7% | 16.8% |

Table 2. Results of different fine-grained models on the test dataset.

| Fusion method | Public | Private |
|---|---|---|
| Fusion method v1 | 14.2% | 14.3% |
| Fusion method v2 | 13.3% | 13.0% |
| **Fusion method v3(final result)** | **12.9%** | **12.9%** |

Table 3. The results of the combination of generic image recognition and fine-grained recognition.

## 3. Result

The following Table 1 lists the training results of some generic image recognition models. We did not submit the results of all models, so the results only listed the results of Res2Net200_vd_26w_4s and multi-model fusion.

The following Table 2 lists some of the training results of fine-grained models. In order to obtain better results, we expanded the resolution to 600x600.

The following Table 3 lists the results of adjusting the fusion thresholds of different models. The fusion v3 method is our final result. Through the meticulous combination of generic and fine-grained models, we got the third place in the Semi-Supervised Recognition Competition of FGVC7.

## 4. Conclusion

In this paper, we introduced our method of mining data, gave some conclusions and some tricks for training generic image recognition models in detail, and at the same time, we introduced methods and conclusions for training fine-grained image recognition models. Also, we described the combination method of the generic image recognition model results fusion and the fine-grained image recognition model results fusion. In the end, we got the third place in the Semi-Supervised Recognition Competition of FGVC7.